\def\year{2026}\relax

\documentclass[letterpaper]{article}
\usepackage{aaai25} 
\usepackage{times} 
\usepackage{helvet} 
\usepackage{courier} 
\usepackage[hyphens]{url} 
\usepackage{graphicx} 
\usepackage{natbib}  
\usepackage{caption}  
\usepackage{algorithm}
\usepackage{algorithmic}

\usepackage{enumitem}
\usepackage{booktabs}
\usepackage{amsmath}
\usepackage{multirow}
\usepackage{dblfloatfix}
\usepackage{threeparttable}
\usepackage{array}

\usepackage{xcolor}

\usepackage{newfloat}
\usepackage{listings}
\floatstyle{ruled}
\newfloat{listing}{tb}{lst}{}
\floatname{listing}{Listing}
\title{TruthStance: An Annotated Dataset of Conversations on Truth Social}
\author{
    Fathima Ameen,
    Danielle Brown,
    Manusha Malgareddy,
    Amanul Haque
}
\affiliations{
    North Carolina State University, Raleigh, NC, USA\\
    fameen@ncsu.edu, mmalgar@ncsu.edu, dmbrow28@ncsu.edu, amanul.003@gmail.com
}

\begin{document}

\maketitle

\begin{abstract}
Argument mining and stance detection are central to understanding how opinions are formed and contested in online discourse. However, most publicly available resources focus on mainstream platforms such as Twitter and Reddit, leaving conversational structure on alt-tech platforms comparatively under-studied. We introduce \textsc{TruthStance}, a large-scale dataset of Truth Social conversation threads spanning 2023--2025, consisting of $24{,}352$ root posts and $523{,}360$ comments with reply-tree structure preserved. We provide a human-annotated benchmark of $1{,}500$ instances across argument mining and claim-based stance detection, including inter-annotator agreement, and use it to evaluate large language model (LLM) prompting strategies. Using the best-performing configuration, we release additional LLM-generated labels for $24{,}352$ posts (argument presence) and $107{,}873$ comments (stance to parent), enabling analysis of stance and argumentation patterns across depth, topics, and users. All code and data are released publicly.
\end{abstract}

\begin{links}
  \link{Code}{https://github.com/MiaAmeen/BlueSocial}
  \link{Dataset}{https://doi.org/10.5281/zenodo.18251738}
\end{links}

\section{Introduction}
Argument mining and stance detection are core tasks in computational discourse analysis, aimed at understanding how opinions are formed, expressed, and contested in text. Argument mining focuses on identifying whether a text contains an argument–typically defined as a \textit{claim} that is supported or challenged by \textit{premises} \cite{SchaeferStede2021}. In contrast, stance detection seeks to determine whether an author expresses support for, opposition to, or neutrality toward a specified target or claim~\cite{SchaeferStede2021}. Unlike argument presence, stance is inherently relational: it must be interpreted with respect to a specific claim or topic.

\begin{figure}[t]
    \centering
    \includegraphics[width=0.47\textwidth]{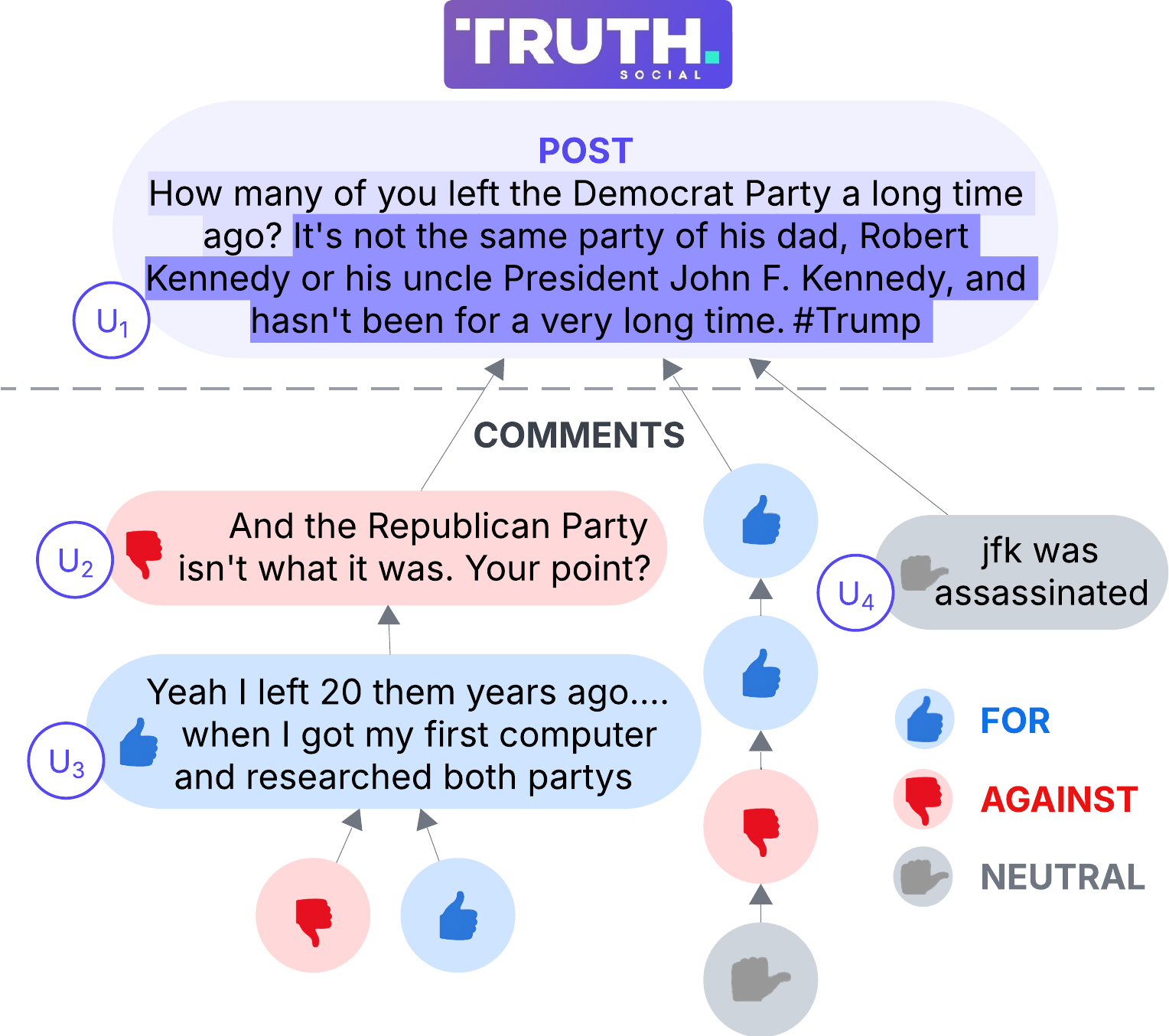}
    \caption{An example conversation tree on Truth Social in which the Original Poster (OP) presents an argument about the U.S. Democratic Party. Commenters express mixed stances in response. The argument claim and premise are highlighted.}
    \label{fig:dialog}
\end{figure}

This relational nature makes \emph{claim-based stance detection} particularly well suited to social media conversations. In this setting, stance can be framed as a sentence-pair classification problem, in which a comment is evaluated relative to the claim advanced by another user–most commonly, the post to which it directly responds. Consequently, many influential stance detection datasets are derived from online discussion platforms such as Twitter and Reddit, from which nested thread-like structures of conversations can be easily extracted \cite{SemEval2017, SRQ2020, ferreira2016emergent}. Figure~\ref{fig:dialog} illustrates this structure: U$_{1}$, the original poster (OP), advances the political claim that the Democratic Party should be abandoned, supported by a premise regarding its perceived decline since the presidency of John F. Kennedy. Subsequent comments to this original post express a range of stances toward this claim. Together, a \textit{post} and its nested \textit{comments} constitute a tree-structured \textit{conversation thread}. Accordingly, parent-child relationships exist both between the original post and its comments, and recursively among comments that respond to earlier comments within the thread.

While the stance of direct replies to the root post can often be inferred straightforwardly, determining the stance of deeper, nested replies is considerably more challenging. For example, in Figure~\ref{fig:dialog}, evaluating U$_{3}$'s comment in isolation may suggest support for the OP’s argument. However, when U$_{2}$'s intermediate reply is taken into account, it becomes clear that the nested comment instead expresses opposition. Prior work has proposed models that explicitly encode full conversational structure to address this challenge~\cite{CSD2023}. While effective, these approaches are often complex, data-intensive, and difficult to scale.

Recent advances in Large Language Models (LLMs) offer a compelling alternative  \cite{PromptFinetuneLLMAM2025}. LLMs have demonstrated strong performance on both argument mining and stance detection. Crucially, LLMs are capable of interpreting local conversational context. If an LLM can reliably infer the stance of a reply relative to its immediate parent, then the stance of any comment with respect to the original post can be inferred by traversing the conversation tree. This observation motivates our approach: we leverage LLMs to perform large-scale, claim-based stance detection over entire conversations, enabling fine-grained analysis of stance dynamics in social media.

Despite substantial progress in stance and argument mining, existing datasets remain heavily skewed toward mainstream platforms. Most widely used resources are overwhelmingly derived from Twitter and Reddit \cite{SemEval2017Task8, SRQ2020}. This narrow platform focus overlooks a rapidly growing segment of the online media ecosystem: \emph{alt-tech} platforms. Alt-tech includes platforms like Gab \cite{GabSocial}, Parler \cite{Parler}, Bluesky \cite{BlueSkyData}, and Truth Social, which emerged in response to perceived ideological bias and moderation practices on mainstream platforms, explicitly positioning themselves as spaces for alternative political discourse and minimal content moderation \cite{altTech}. Although these platforms host smaller user bases, research increasingly suggests that they constitute a parallel media system that both reacts to and influences mainstream political communication \cite{GabSocial_polarization}. Truth Social, in particular, has played a visible role in contemporary political discourse \cite{TrumpTwitterTruthSocial}. Yet, despite its relevance, \emph{conversations on Truth Social remain largely unexplored in the literature}. This gap is due in part to data limitations. Existing Truth Social datasets primarily consist of isolated posts and lack conversational context \cite{gerard2023truthsocial, TS24}. Without reply structure, it is impossible to study dialogical phenomena such as disagreement, persuasion, or stance evolution–processes that are central to understanding political argumentation. To the best of our knowledge, \emph{no publicly available dataset captures large-scale conversation threads on Truth Social}.

In this work, we address this deficit by extending an existing Truth Social post-level dataset, and collecting and releasing a large-scale corpus of approximately 24K conversation threads, comprising over 523K newly scraped comments from 2023–2025, with full post-comment structure preserved. This dataset enables the first systematic study of argumentation on Truth Social. Leveraging LLMs, we annotate (i) original posts for argumentative content and (ii) comments for claim-based stance relative to their parent posts, allowing us to track how arguments and stances evolve across conversation depth. Finally, we conduct quantitative and qualitative analyses of argumentation and stance dynamics, providing new empirical insights into political discourse on an under-studied alt-tech platform.

\paragraph{Contributions.}
\begin{enumerate}
    \item We release \textsc{TruthStance}, a dataset of $24{,}352$ Truth Social conversation threads (2023--2025) containing $523{,}360$ comments with full reply structure, along with associated post- and author-level metadata.
    \item We provide $1{,}500$ ground-truth labels across argument mining and claim-based stance detection (with inter-annotator agreement) and use this set to evaluate LLM prompting strategies and a supervised baseline.
    \item Using the best-performing configuration, we release $24{,}352$ LLM annotations for argument presence and $107{,}873$ LLM annotations for stance-to-parent, and present initial analyses of argumentation and stance expression across topics, conversation depth, and users.
\end{enumerate}



\subsection{LLMs for Argument Mining and Stance Detection}

Plenty of prior work has applied argument mining and stance detection to social media data \cite{AldayelReview2021}. Early claim-based stance datasets such as Emergent~\cite{ferreira2016emergent} annotate tweets with respect to news headlines; however, posts are provided in isolation without any conversational context, limiting the ability to model stance in an interactive setting. Subsequent work has emphasized the importance of conversational structure. Datasets such as SRQ~\cite{SRQ2020}, Cantonese-CSD~\cite{CSD2023}, and MT-CSD~\cite{MT-CSD} incorporate conversation threads from platforms including Twitter, Hong Kong social media, and Reddit respectively. These resources provide richer context, but they frame stance as \emph{target-based}, restricting annotations to predefined entities or topics rather than explicit claims articulated within the conversation. The dataset most closely aligned with our setting is SemEval-2017 Task 8 (RumourEval)~\cite{SemEval2017}, which provides claim-based stance annotations within conversation threads on Twitter. Replies are labeled as \textit{Support}, \textit{Deny}, \textit{Query}, or \textit{Comment} with respect to a rumor introduced at the root of the conversation. While this dataset establishes an important precedent for conversational, claim-based stance detection, it is limited to Twitter and focuses primarily on rumor verification rather than general argumentative discourse. To our knowledge, no existing dataset combines claim-based stance and argument annotation and full conversational structure on alt-tech media.

Early approaches to stance classification relied on traditional supervised models, including support vector machines, logistic regression, decision trees, and k-nearest neighbors, with SVMs being particularly prevalent~\cite{aker2017kNNRumourEval}. Subsequent work introduced neural architectures that explicitly encode conversational context. For example, Poddar et al.~\cite{Poddar2018ctstance} combined CNN-based tweet encoders with RNNs and attention mechanisms to model conversational flow, achieving strong performance on stance benchmarks. Branch-LSTM~\cite{Branch-LSTM} further incorporated conversation structure by processing entire reply branches using LSTM units~\cite{LSTM}. More recent efforts leverage pretrained language models: \cite{CSD2023} introduced CNN-based models over BERT embeddings, later extending this architecture with graph convolutional networks to encode reply structure more explicitly~\cite{MT-CSD}. Despite these advances, supervised stance models trained on existing benchmarks often exhibit poor out-of-domain generalization~\cite{Ng2022StanceCrossValidation}. This limitation is especially pronounced in our setting, as Truth Social represents a niche, ideologically homogeneous platform whose discourse differs substantially from mainstream platforms such as Twitter and Reddit. Consequently, models trained on prior datasets may not transfer reliably to this domain.

Large language models (LLMs) have recently emerged as a promising solution to this challenge. Surveys demonstrate that LLMs substantially advance argument mining and stance detection through zero-shot and few-shot learning, scalable annotation, and even dataset synthesis~\cite{li2025LLM_AM_survey, ColaSD_Lan2024, SD_Mets2024, GPTSD_Liyanage2023, SDSM_li2023, SentimentGPT2023, GPTSD_Aiyappa2023, Yuan2025Benchmark}. In-context learning techniques such as few-shot prompting~\cite{brown2020language} and chain-of-thought reasoning~\cite{wei2022chain} have been shown to yield competitive or state-of-the-art performance on stance benchmarks including SemEval-2016 and P-Stance~\cite{Zhang2024StanceDetectionGPT, zhang2024investigatingchainofthoughtchatgptstance}. Additional work has explored LLM-based rationale generation and distillation to supervise smaller models~\cite{Yuan2025Reasoner}, further reducing annotation costs. While political bias in LLM-based stance classification has been observed, these effects primarily arise at the dataset level and can be mitigated through consistent prompting strategies~\cite{LLM-SD-political-bias}.

However, existing LLM-based approaches predominantly evaluate each post independently or relative only to a single target or root claim. To our knowledge, no prior work systematically leverages LLMs to traverse conversation threads and iteratively infer stance along parent-child reply chains. This gap is particularly salient for deeply nested discussions, where stance cannot be reliably inferred without considering intermediate replies. Our work addresses this limitation by explicitly modeling stance propagation across conversational structure using LLM-based annotation.

\subsection{Conversations on Truth Social}

Research on Truth Social remains scarce, with most prior work focusing on data collection rather than discourse analysis. Gérard et al.~\cite{gerard2023truthsocial} introduced the first publicly available dataset from Truth Social, collected in 2023, followed by an expanded release covering 2025~\cite{TS24}. These datasets, however, consist exclusively of isolated posts and do not provide access to full conversational threads.

A rare exception is the study by \cite{wikiTruths}, which examined the presence of Wikipedia links in Truth Social posts. The authors found that posts containing Wikipedia references consistently received lower engagement than posts without such links, suggesting that the neutral tone of Wikipedia-linked content may reduce the likelihood of eliciting responses or debate. Beyond this analysis, however, there is a notable absence of research on discourse and argumentative interactions on Truth Social. This gap motivates the collection and analysis of complete conversational threads, enabling the study of stance and argumentation in a politically homogeneous, alternative social media context.


\section{Methodology}

\begin{table*}[t]
\centering
\begin{threeparttable}
\caption{Summary statistics of the Truth Social dataset across preprocessing stages.}
\label{tab:descriptive}
\begin{tabular}{lccccccccc}
\hline
\textbf{Group} &
\textbf{\#Posts} &
\textbf{\#Users} &
\textbf{Likes$_{\text{avg}}$} &
\textbf{Replies$_{\text{avg}}$} &
\textbf{ReT$_{\text{avg}}$} &
\textbf{Fwr.$_{\text{avg}}$} &
\textbf{Fwg.$_{\text{avg}}$} &
\textbf{Length$_{\text{avg}}$} &
\textbf{Depth$_{\text{avg}}$} \\
\hline
Raw Data  &$776{,}281$&$39{,}446$&$5.86$&$0.61$&$1.99$& \textbf{NA} & \textbf{NA} & 226 & \textbf{NA} \\
\hline
Arg. Posts       &$12{,}271$&$2{,}731$&$96.50$ &$20.76$&$44.54$&$89.5$K &$6.2$K & 286 & 3.11 \\
Non-arg. Posts   &$12{,}081$&$2{,}143$&$155.94$&$25.69$&$56.24$&$278.8$K&$6.2$K & 166 & 3.14 \\
\hline
Comments$_{\text{AGAINST}}$ &$18{,}995$&$8{,}486$&$1.42$&$0.73$&NA&$1.8$K&$1.2$K & 163 & 1.76 \\
Comments$_{\text{FOR}}$     &$64{,}238$&$24{,}475$&$3.35$&$0.36$&NA&$3.2$K&$2.2$K & 115 & 1.30 \\
Comments$_{\text{NEUTRAL}}$ &$24{,}640$&$10{,}839$&$1.99$&$0.49$&NA&$3.8$K&$2.3$K & 88 & 1.35 \\
\hline
\end{tabular}
\begin{tablenotes}
\footnotesize
\item Average values are computed per post. ReT refers to the count of retruths. Fwr. and Fwg. refer to the average follower and following counts, respectively, of post authors. Length refers to the average number of characters per post/comment after preprocessing. Depth$_{avg}$ refers to the average reply depth of the associated conversation tree. For comments, Depth$_{avg}$ is computed as the mean thread depth of all conversations originating from \textit{direct replies} (first level comments). NA indicates unavailable metadata.
\end{tablenotes}
\end{threeparttable}
\end{table*}
In this section, we describe the steps taken for collection, pre-processing, and augmentation of our conversational Truth Social dataset. We then outline the procedures employed for argument mining and stance detection, followed by a description of the additional structural and engagement-based metrics used in our analyses.

\subsection{Datasets}
Our dataset builds on the publicly available Truth Social dataset introduced by \cite{TS24}, which contains approximately 776K posts collected between February and October 2024, and is licensed for reuse with attribution for non-commercial purposes (\url{https://creativecommons.org/licenses/by-nc/4.0/}). The dataset focuses on political discourse and was compiled using a hybrid data collection strategy. First, posts associated with daily trending political hashtags (e.g., \textit{trump2024}) were scraped. Second, a predefined set of politically salient keywords was continuously monitored, and posts containing these keywords were collected in the same manner as hashtag-based posts. Together, these strategies provide broad coverage of politically relevant content on the platform. To prepare the data for analysis, we applied the following preprocessing steps:
\begin{enumerate}
    \item Removed posts with missing or empty textual fields.
    \item Deduplicated posts with identical (\textit{author, text}) pairs, which are indicative of automated or spam-like behavior.
    \item Removed reposts (``retruths''), as they do not contain original user-generated content.
    \item Excluded posts with fewer than three direct replies to ensure the presence of meaningful conversational interaction.
    \item Removed posts lacking substantive textual content. Text was considered substantive if, after removing hashtags and URLs via regular expressions, the remaining content was non-empty.
\end{enumerate}

After applying these pre-processing steps, $24,378$ posts remained and were included in the analysis. The original dataset includes post-level metadata such as author identifiers, textual content, timestamps, and engagement statistics (likes, replies, and retruths). However, while the original dataset provides reply counts, it does not provide access to the parent comments or root posts themselves. Consequently, complete conversation threads cannot be reconstructed from the original dataset alone. In addition, the dataset does not contain author-level metadata (e.g., follower counts, following counts, or user bios), which limits analyses of user-level and social-contextual factors.

\subsubsection{Dataset Augmentation and Enrichment}
To address these limitations, we augmented the dataset using the Truth Social API, which constitutes our first contribution. Using a custom scraping tool built on Stanford’s open-source social media collection framework \cite{McCain_Truthbrush_2022}, we retrieved the full set of comments associated with all filtered posts in the preprocessed dataset. Each comment includes a unique identifier ($id$) and an $in\_reply\_to\_id$ field, which allows us to computationally define a conversation thread as a directed acyclic tree rooted at a single post, with edges corresponding to $in\_reply\_to\_id$ relationships.

In parallel, we enriched both original posts and newly collected comments with author-level metadata obtained directly from the platform, including follower and following counts. Table~\ref{tab:descriptive} reports summary statistics across pre-processing and enrichment stages, including the total number of posts, unique users, and average engagement metrics per post. For the enriched dataset, we additionally report the average follower and following counts per user. In total, we collected $523{,}360$ comments corresponding to $89{,}466$ unique commenters, and $3{,}886$ author records for posters, resulting in $90{,}593$ unique authors overall.

\subsection{Annotation Task Definition}

\begin{table*}[t]
\centering
\caption{Performance of LLMs and baseline SVM on Argument Mining (binary) and Stance Detection (3-class). Entries are macro-F1 / accuracy. Bold indicates the best result per row. McNemar $p$-values indicate significance of pairwise prompting differences within each LLM ($^*$ $p<0.05$, $^{**}$ $p<0.01$, $^{***}$ $p<0.001$).}
\label{tab:llm_results_updated}
\begin{tabular}{l|ccc|ccc}
\hline
\multirow{2}{*}{Model} & \multicolumn{3}{c|}{Argument Mining} & \multicolumn{3}{c}{Stance Detection} \\
\cline{2-7}
 & CoT & Few-shot & CoT+Few-shot & CoT & Few-shot & CoT+Few-shot \\
\hline
\textbf{LLMs} & & & & & & \\
gemini-2.5-flash & 0.806 / 0.808 & 0.816 / 0.819 & 0.821 / 0.822 & 0.706 / 0.786 & 0.622 / 0.714 & \textbf{0.783 / 0.836}$^{***}$\\
DeepSeek-v3       & 0.821 / 0.823 & 0.819 / 0.826 & 0.822 / 0.823 & 0.642 / 0.669 & 0.674 / 0.686 & 0.677 / 0.702 \\
GPT-o3            & \textbf{0.821 / 0.826}$^{***}$ & 0.712 / 0.712 & \textbf{0.826 / 0.830}$^{***}$ & \textbf{0.671 / 0.721}$^{***}$ & 0.326 / 0.415 & 0.318 / 0.424 \\
\hline
\textbf{Baseline} & & & & & & \\
SVM & 0.757 / 0.763 & - & - & 0.674 / 0.769 & - & - \\
\hline
\end{tabular}
\end{table*}

Our annotation pipeline proceeds in two stages. First, we identify argumentative posts via argument mining. Second, for posts labeled as argumentative, we perform claim-based stance detection on all replies within the conversation thread.

\subsubsection{Argument Mining}

We frame argument mining as a binary classification task: determining whether a post contains a claim supported by at least one premise. For ground truth, two authors independently annotated an initial random sample of 100 posts, labeling each as argumentative or non-argumentative. This procedure was repeated for a second batch of 100 posts, yielding a Cohen’s $\kappa$ of 0.70, corresponding to substantial inter-annotator agreement according to \cite{LandisKoch}. Disagreements were resolved through discussion, after which an additional 550 posts were annotated, resulting in a total of $750$ human-labeled posts for training and evaluation. Of the $750$ posts, $326$ were identified as argumentative and $425$ as non-argumentative, resulting in a roughly balanced class distribution. Appendix A~\ref{app:annotations} contains a full description of the annotation guidelines followed by authors for the task.

\begin{figure}[H]
    \centering
    \includegraphics[width=0.45\textwidth]{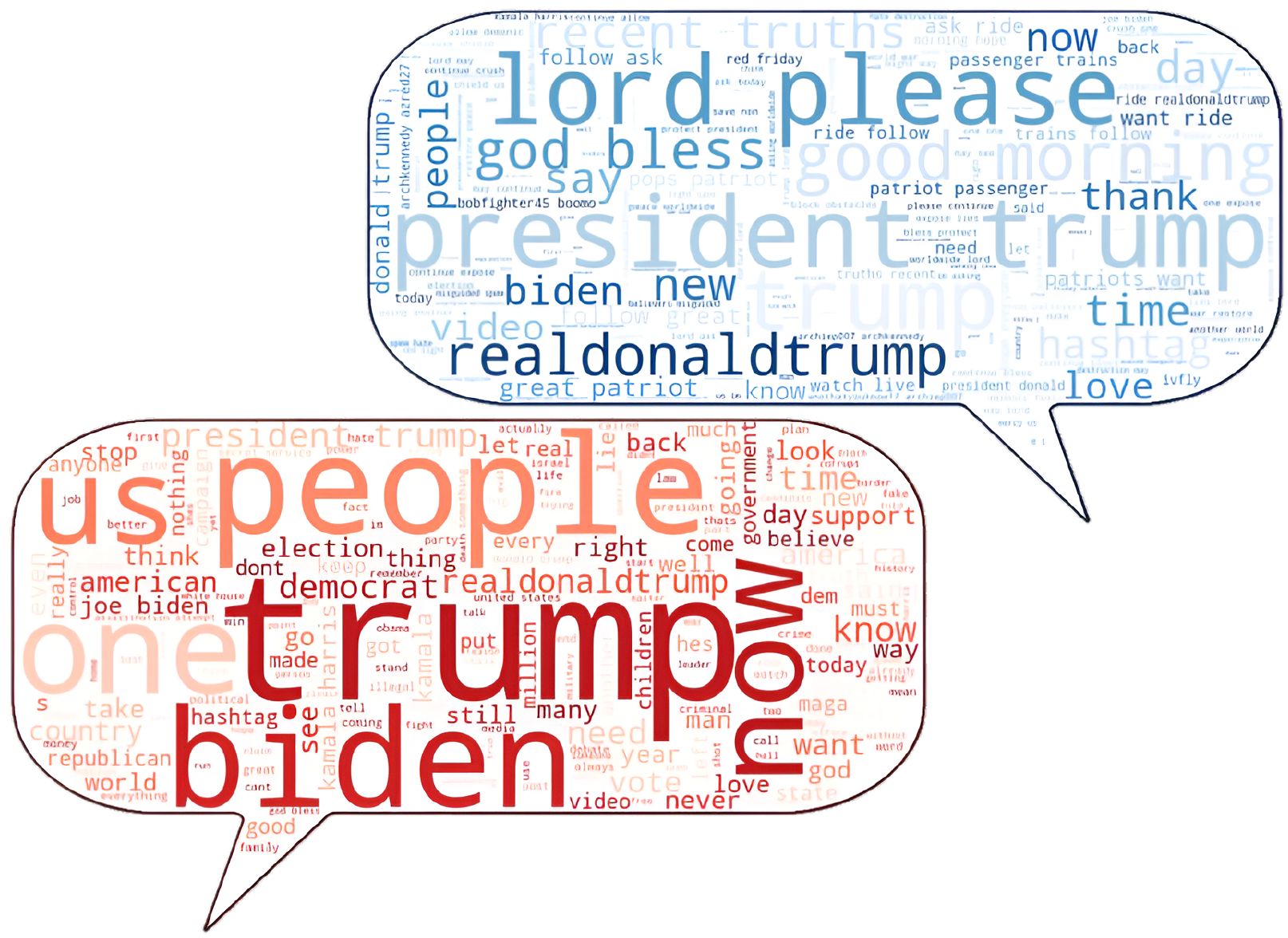}
    \caption{Wordclouds of argumentative and non-argumentative posts on Truth Social.}
    \label{fig:wordcloud}
\end{figure}

Using this subset, we evaluated several LLM-based classifiers under different prompting strategies, selecting the best-performing model for annotating the remainder of the dataset. The next section provides full details of the LLM annotation setup. The selected model produced annotations for a $24{,}352$ of the clean posts. Figure~\ref{fig:wordcloud} visualizes word frequency distributions across argumentative and non-argumentative posts. Non-argumentative posts often contain prayers (e.g., ``lord please", ``god bless"), whereas argumentative posts frequently reference political actors such as ``Biden" and action-oriented terms such as ``people", ``us", and ``now", suggesting calls to action.

\subsubsection{Stance Detection}
Our second goal is to track how stances propagate through conversational threads initiated by argumentative posts. We frame stance detection as a sentence pair classification task; determining whether a comment is in favor of, opposed to, or neutral to a claim expressed by its immediate parent (comment or post). Because a conversation thread is nested, we can recursively traverse parent–child relations to also infer any comment's stance toward the OP's argumentative post. Each comment is annotated relative to its parent as one of three labels:

\begin{itemize}
    \item \textbf{FOR}: the comment agrees with or reinforces the parent’s claim.
    \item \textbf{AGAINST}: the comment disputes or challenges the parent’s claim.
    \item \textbf{NEUTRAL}: the comment does not express a clear stance toward the parent’s claim.
\end{itemize}
Appendix A~\ref{app:prompts} includes the annotation guidelines for this task as well. We note an important limitation of our approach to inferring child stance to the OP: because NEUTRAL denotes the absence of a clear stance relation to the parent, stance cannot be reliably propagated through a chain containing a neutral intermediate reply. In such cases, the stances of child comments branching from a neutral comment become undefined. We therefore exclude all descendant branches rooted at any NEUTRAL comment from OP-level stance analyses.

We followed an annotation procedure analogous to argument mining, using stance-specific guidelines. Two coders independently annotated an initial subset of comments, achieving substantial agreement (Cohen’s $\kappa=0.76$), after which disagreements were resolved through discussion. The coders then annotated an additional $750$ comments, of which $452$, $168$, $132$ belong to the FOR, AGAINST, and NEUTRAL classes, respectively, yielding a labeled dataset for downstream modeling and analysis. 

We next selected the LLM with the highest performance on this subset to annotate the remainder of the dataset; we detail this selection and annotation process in the next section. Due to time and resource constraints, the LLM was not applied to the full corpus of $523K$ comments; instead, we obtained approximately $107{,}873$ LLM-generated annotations for comments of randomly selected conversation threads, which complement the ground-truth labels. Table~\ref{tab:descriptive} reports the final dataset sizes and label distributions, including both human-annotated and LLM-annotated data.

\subsection{Annotation Pipeline} \label{sec:LLM_annotation_pipeline}
For both argument mining (AM) and stance detection (SD), we designed a standardized prompt template consisting of a task definition and a required output format. The task definition ensures that the LLM understands the objective, while the output format constrains responses to a set of predefined labels. For AM, the prompt consists of a single post and a request for a binary label (argumentative or non-argumentative). For SD, the prompt includes a post-comment pair and requests a multi-class label corresponding to the stance of the comment. All prompts are made publicly available in our GitHub repository; we also include the specific instructions used for each template in Appendix B~\ref{app:prompts}. We opted against fine-tuning LLMs for the task, as prior work \cite{PromptFinetuneLLMAM2025} found that finetuning did not improve model performance for the stance detection task and in fact, degraded it. This may be because finetuning makes models too specialized, thereby impairing out-of-domain generalization \cite{FineTuningBad2022}. Consequently, we focus on few-shot LLM inference, which is more robust to domain shifts.

To assess performance, we compared several LLMs with a baseline support vector machine (SVM) model, which has historically performed well on SemEval-2016 stance detection tasks \cite{SVM_SemEval2016}. For the SVM, we used \texttt{Qwen/Qwen3-Embedding-0.6B} \cite{Qwen3Embeddings} to generate sentence embeddings, as it is both lightweight and high-performing relative to alternative embeddings \cite{LLMARENA}. For SD, embeddings of both the comment and its immediate parent post (or comment) were concatenated to provide as input, whereas for AM, only the post embedding was used. We used stratified 5-fold cross-validation and generated out-of-fold predictions for every example. Specifically, each item was assigned a final label based on the prediction from the fold in which it was held out from training.

We evaluated three LLMs: \textbf{Gemini-2.5-Flash} \cite{gemini}, \textbf{GPT-o3}, and the open-source \textbf{DeepSeek-v3} \cite{DSv3}, which have knowledge cutoff dates of January 2025, June 2024, and July 2024, respectively. They were accessed via their official API endpoints. We tested three prompting strategies: Few-Shot (FS), Chain-of-Thought (CoT), and CoT+Few-Shot (FSCoT)–each prompting instruction can be accessed in Appendix B~\ref{app:prompts}. For each configuration, predictions were aggregated using a majority vote over three independent runs. Performance was measured using macro-F1 and accuracy. Additionally, we conducted McNemar's test to assess the statistical significance of differences between (1) prompting strategies within each LLM and (2) LLM within each prompting strategy. Results are summarized in Table~\ref{tab:llm_results_updated}.

\paragraph{Interpretation of Results.} 
For \textbf{AM}, all models exhibit comparable performance, with no consistent winner across prompting strategies. Gemini and DeepSeek perform similarly under FS, CoT, and FSCoT prompting, while GPT shows equivalent performance under CoT and FSCoT. LLMs generally outperform the SVM baseline on AM, although performance varies with prompting strategy.

For \textbf{SD}, the task is more challenging. DeepSeek exhibits minimal differences across prompting strategies. Gemini performs best under FSCoT, while GPT achieves its peak with CoT. Overall, Gemini with FSCoT achieves the highest macro-F1 and accuracy, outperforming both the other LLMs and the SVM baseline. Consequently, we adopt \textbf{Gemini with FSCoT} for stance detection. For argument mining, given similar performance across prompts, we also choose \textbf{Gemini} for ease of implementation.

\subsection{Conversational Features}
To characterize how arguments are received and how stance evolves within conversations, we extract a set of content-level, engagement-level, and user-level features for each conversation thread. These features complement our argument mining and stance classification outputs and enable a multi-dimensional analysis of discourse dynamics on Truth Social.

\subsubsection{Post Content features.} To capture emotional tone, we apply the lexicon-based VADER sentiment analyzer \cite{hutto2014vader}, which has been widely used in social media analysis and stance prediction. We further assess the prevalence of hostile or abusive language using Detoxify \cite{hanu2020detoxify}, an open-source neural model trained for multi-label toxic comment classification. Detoxify outputs continuous scores in $[0,1]$ for toxicity. These scores allow us to quantify the intensity of discourse and examine how toxicity correlates with argumentative behavior and stance.
As a proxy for deliberative effort, we measure post length after normalizing text by removing URLs, hashtags, emojis, and markup. We compute character counts on the cleaned text and report average lengths by stance and argument category. As reference, an \texttt{X} post is currently limited to $280$ characters. We argue that the higher number of characters in the posts indicates more effort, which possibly translates to more substantiated opinions.

\subsubsection{Post engagement metrics.} In addition to textual content, we leverage platform-provided engagement metadata for posts and comments. For each post, we record the number of likes, replies, and re-shares (``retruths"). While re-share data is unavailable for comments, we can still assess how argumentative content is received and amplified at the post level.
Additionally, we compute the maximum reply depth of each conversation, defined as the longest path from the root post to a terminal comment (a comment that was not replied to). Deeper threads may reflect sustained engagement or prolonged disagreement around an argument.

\subsubsection{User Features} Our dataset includes user metadata for all posts and comments, enabling analysis of speaker characteristics. For each user, we extract follower count, following count, verification status (binary), and profile bio text. Verification status is treated as a proxy for institutional or platform-recognized prominence. 

\subsubsection{Topic Modeling}
In our dataset, we found $28{,}582$ unique hashtags used in posts and comments. Hashtags serve as a sufficiently informative proxy for the topical content of a post \cite{hashtagstopicmodeling}. Therefore, we performed topic modeling by clustering semantically similar hashtags. We again used the \texttt{Qwen/Qwen3-Embedding-0.6B} embedding model to generate dense representations of the hashtags. We then applied HDBSCAN, a density-based clustering algorithm, to group hashtags based on their semantic similarity in the embedding space \cite{hdbscan}. HDBSCAN was selected for its ability to discover clusters of varying densities and to assign noise labels to hashtags that do not belong to any coherent topic cluster. We set the minimum cluster size to $20$ and the minimum number of samples to $3$ to balance topic granularity and robustness. The resulting clusters represent emergent topical groupings of hashtags, with unassigned hashtags treated as outliers. The author assigned labels to each cluster after manual inspection. Table~\ref{tab:topics_hashtags} shows the top-mentioned topics in the dataset along with example hashtags associated with each topic.

\begin{table}[h]
\centering
\caption{Top mentioned topics and example hashtags.}
\label{tab:topics_hashtags}
\begin{tabular}{>{\bfseries}l >{\raggedright\arraybackslash}p{6cm}}
\toprule
Topic & Hashtags \\
\midrule
maga & \#maga4life, \#trumpmaga, \#magagop, \#freedonaldtrump, \#2024trumpvance \\
trump & \#trump, \#trumpnews, \#pesidenttrump, \#trumpism, \#realdonaldjtrump\\
biden & \#biden2024, \#bidenadmin, \#bidenisdone \\
democrats & \#democrathate, \#obamaadmin, \#hilaryclinton, \#democratelites \\
deception & \#realitycheck, \#truthberevealed, \#seekthetruth, \#thetruthwillsetyoufree \\
\bottomrule
\end{tabular}
\end{table}

\section{Descriptive Overview}
\begin{figure*}[b]
    \centering
    \includegraphics[width=\textwidth]{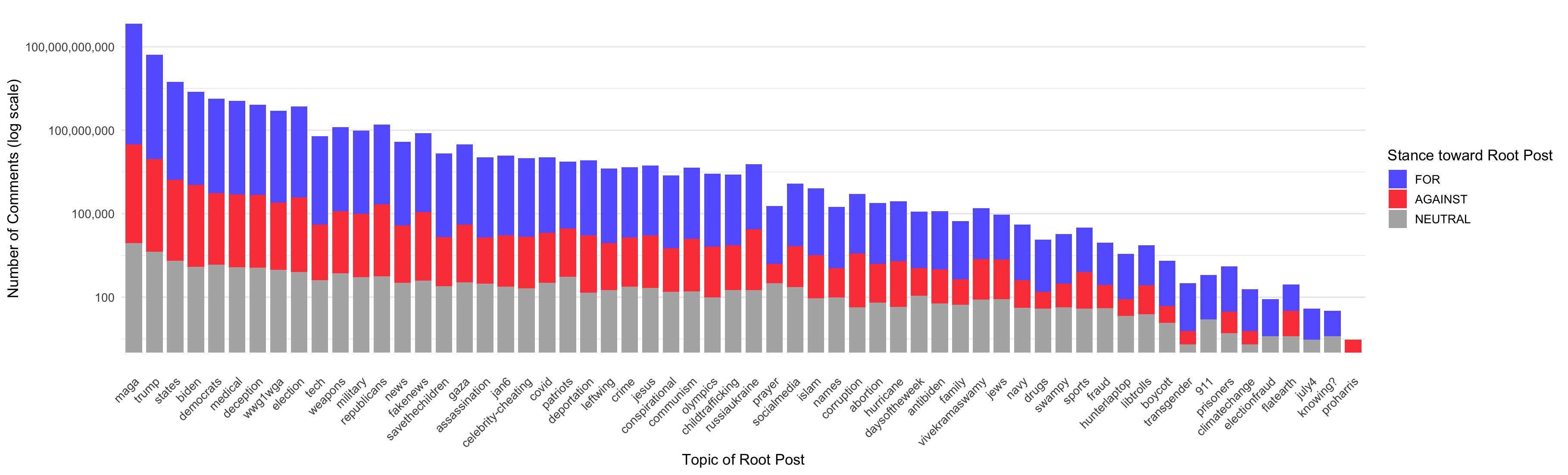}
    \caption{Stacked bar plot showing the most frequently discussed topics in argumentative posts, along with the distribution of comment stances for each topic. Topics are ordered by total comment volume.}
    \label{fig:bargraph}
\end{figure*}
We provide a descriptive characterization of conversations in our dataset by qualitatively examining argumentative posts and analyzing how stance is distributed–and evolves–throughout conversations.

\subsection{Characterizing an Argument} To examine predictors of argumentative posts, we fitted a series of nested logistic regression models. The baseline model included only an intercept. Subsequent models sequentially added blocks of the predictors previously discussed: engagement metrics, post content features, and author traits. Logistic regression was implemented using the \texttt{glm} function in R with a binomial family, and model comparisons were conducted via analysis of deviance (likelihood ratio tests) to assess the incremental contribution of each predictor block.

\begin{table}[h]
\centering
\begin{threeparttable}
\caption{Logistic regression odds ratios predicting AM labeling.}
\begin{tabular}{lcc}
\hline
\textbf{Predictor} & \textbf{Odds Ratio} & \textbf{95\% CI} \\
\hline
\multicolumn{3}{l}{\textit{Intercept}} \\
Intercept & \textbf{0.69}$^{***}$ & [0.66, 0.73] \\
\hline
\multicolumn{3}{l}{\textit{Post Engagement}} \\
Like count & \textbf{0.999}$^{***}$ & [0.999, 0.999] \\
Reply count & \textbf{0.995}$^{***}$ & [0.992, 0.997] \\
Retruth count & \textbf{1.004}$^{***}$ & [1.003, 1.005] \\
Max reply depth & 0.998 & [0.991, 1.003] \\
\hline
\multicolumn{3}{l}{\textit{Post Content}} \\
Sentiment & \textbf{0.45}$^{***}$ & [0.42, 0.47] \\
Toxicity & \textbf{5.02}$^{***}$ & [4.22, 5.99] \\
Length & \textbf{1.002}$^{***}$ & [1.002, 1.002] \\
\hline
\multicolumn{3}{l}{\textit{User Traits}} \\
Followers & \textbf{1.000}$^{***}$ & [1.000, 1.000] \\
Following & \textbf{0.999}$^{***}$ & [0.999, 0.999] \\
Verified & 1.05 & [0.98, 1.12] \\
\hline
\label{tab:regression_coeffs}
\end{tabular}
\begin{tablenotes}
\footnotesize
\item Logistic regression odds ratios predicting the likelihood that a post is labeled as argumentative (1 = argument, 0 = not argument). Odds ratios are reported for each predictor, with 95\% confidence intervals in brackets. Bold indicates statistically significant coefficients at $p < 0.001$ (***). Continuous predictors were entered untransformed.
\end{tablenotes}
\end{threeparttable}
\end{table}

Analysis of deviance comparing nested logistic regression models indicates that all blocks of predictors significantly improve model fit. Adding engagement metrics (likes, replies, retruths, and max reply depth) to the intercept-only model resulted in a modest but significant reduction in deviance ($\Delta\text{Deviance} = 303.8, p < 0.001$). Incorporating textual features (sentiment, toxicity, and post length) produced a substantially larger improvement ($\Delta\text{Deviance} = 2{,}468.2, p < 0.001$), suggesting that content is the strongest predictor of AM labeling. Finally, including author-level features (followers, following, and verified status) further reduced deviance ($\Delta\text{Deviance} = 182.2, p < 0.001$), providing additional but smaller predictive value. 

\begin{figure}[t]
    \centering
    \includegraphics[width=0.45\textwidth]{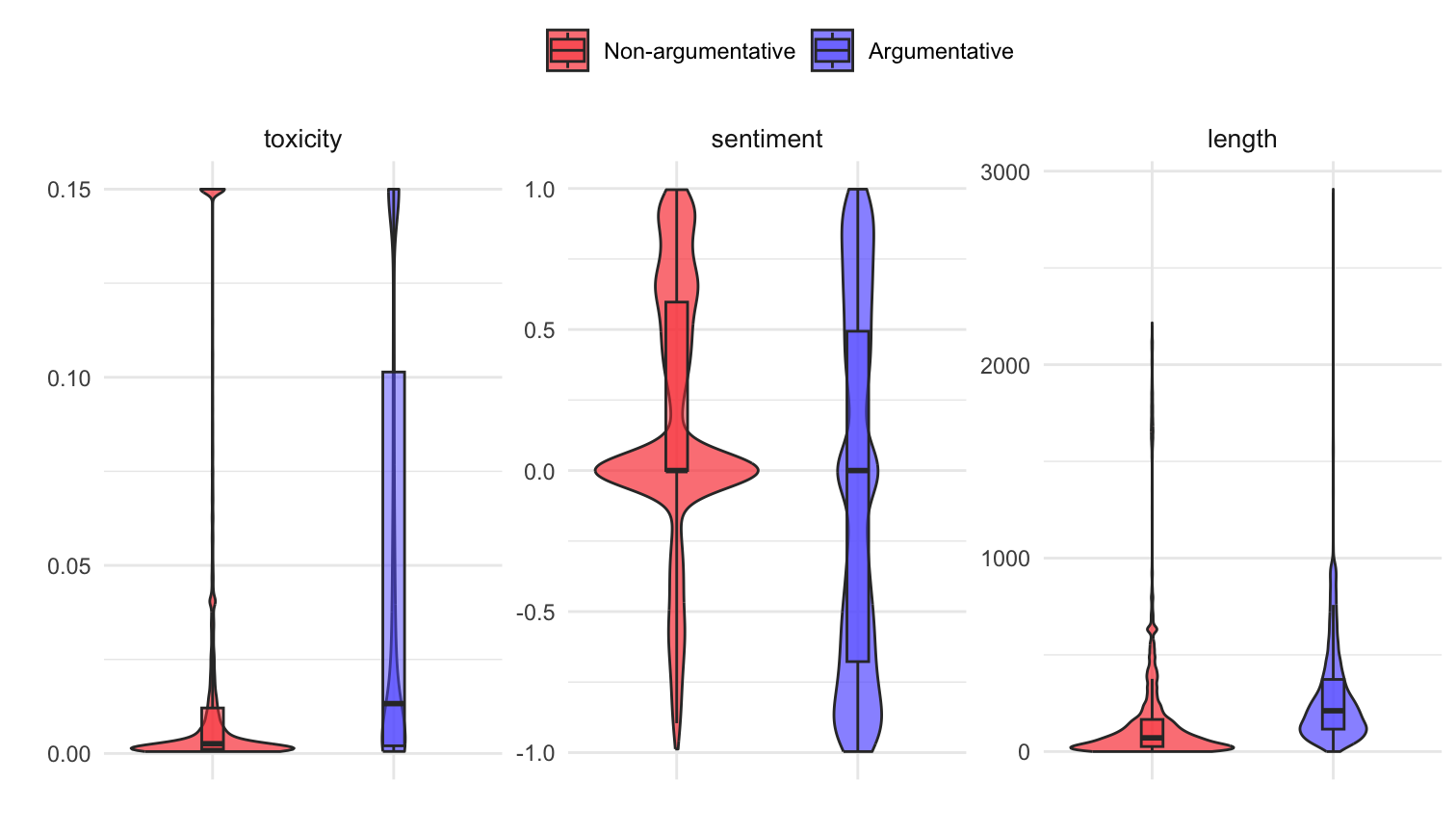}
    \caption{Violin plots comparing the distributions of toxicity, sentiment, and post length for argumentative versus non-argumentative posts. Each panel shows a violin density with an overlaid boxplot summarizing the median and interquartile range.}
    \label{fig:violin}
\end{figure}

The regression results are reported in Table~\ref{tab:regression_coeffs}. The logistic regression highlights several significant predictors of AM labeling. Posts with higher sentiment scores are substantially less likely to be labeled as argumentative, whereas posts with higher toxicity are much more likely to receive an AM label. Engagement metrics show mixed effects: retruth counts slightly increase the odds of argument labeling, while likes and replies are associated with marginal decreases. Importantly, although these engagement effects are statistically significant, their effect sizes are very small, which is expected given that a change of a single like, retruth, or reply is unlikely to meaningfully alter whether a post is argumentative. In terms of user characteristics, users with larger follower counts and smaller following counts are more likely to produce argumentative content. Overall, content-based features–particularly sentiment and toxicity–emerge as the strongest indicators of argument presence. This pattern is further reflected in the distributional comparisons in Figure~\ref{fig:violin}: non-argumentative posts tend to exhibit lower toxicity, sentiment scores closer to neutral (0), and shorter length, whereas argumentative posts span a wider and higher range of toxicity values, include more negatively valenced sentiment, and are generally longer on average.

\subsection{Evolution of stance}
We track stance in the dataset at multiple levels, examining how stance varies across topics, evolves within conversations, and differs across users based on individual stance shifts.

\subsubsection{Topic-level}
We analyze the distribution of comment-level stance across topical categories of argumentative root posts. Comments are aggregated by topic and stance (FOR, AGAINST, NEUTRAL) and visualized using stacked bar charts in Figure~\ref{fig:bargraph}, with topics ordered by total comment volume. To account for substantial variation in engagement across topics, comment counts are shown on a logarithmic scale. The topic receiving the highest volume of comments is \textit{MAGA} (``Make America Great Again''), a slogan closely associated with Donald Trump’s political movement. Other high-engagement topics include \textit{Trump}, \textit{Biden}, and \textit{Democrats}, reflecting the explicitly political focus of the dataset. Despite large differences in overall comment volume, the relative distribution of supportive, opposing, and neutral stances appears broadly consistent across topics, suggesting that stance polarization is not driven by topic alone but is a general characteristic of political discussion on the platform.

\subsubsection{Conversation-level}
Table~\ref{tab:stance_depth} reports, for each reply depth, the proportion and absolute number of replies expressing supportive (FOR), opposing (AGAINST), and neutral (NEUTRAL) stances toward the root post. This analysis provides an initial view of how stance expression varies as conversations progress deeper into comment threads, offering the first large-scale examination of stance on Truth Social.

Across increasing reply depths, we observe a gradual increase in the relative prevalence of neutral responses, accompanied by a corresponding decline in explicit supportive or opposing stances. This pattern suggests that as conversations unfold, participants may be less likely to directly engage with the original argumentative position, potentially reflecting topic drift, reduced engagement with the root claim, or a shift toward conversational maintenance rather than stance-taking.

\begin{table}[t]
\centering
\caption{Distribution of comment stances (\%) and number of (child) comments (\#) to comments across reply depth (D).}
\label{tab:stance_depth}
\begin{tabular}{lcccccc}
\toprule
 & \multicolumn{2}{c}{AGAINST} & \multicolumn{2}{c}{FOR} & \multicolumn{2}{c}{NEUTRAL} \\
\cmidrule(lr){2-3} \cmidrule(lr){4-5} \cmidrule(lr){6-7}
D & \% & \# & \% & \# & \% & \# \\
\midrule
1 & 12.3\% & 306 & 71.8\% & 1K  & 16.0\% & 300  \\
2 & 10.9\% & 89 & 57.7\% & 476 & 31.4\% & 244  \\
3 & 17.9\% & 56 & 36.0\% & 166 & 46.1\% & 185 \\
4 & 11.4\% & 36  &33.8\% & 75  & 54.7\% & 146  \\
5 & 16.5\% & 23  & 23.4\% & 37  & 60.1\% & 95  \\
6+ & 8.8\% & 58  & 12.3\% & 138  & 78.8\% & 517  \\
\bottomrule
\end{tabular}
\end{table}

\subsubsection{User-level}
We focus on conversations in which individual users contribute at least five times at different depths, which we term ``deliberative'' conversations due to the high level of user engagement. For these conversations, we track the evolution of each user’s stance across their turns. We visualize these transitions using a Sankey diagram, where nodes represent stance categories at each turn and links indicate the flow of users between stances across consecutive turns.

The diagram largely reinforces the trends observed in Table~\ref{tab:stance_depth}: users initially expressing a supportive stance (FOR) often shift toward neutral (NEUTRAL) over successive turns, whereas users initially expressing opposition (AGAINST) generally maintain that stance, with relatively few switching. This suggests asymmetric dynamics in how users update or maintain their stance within extended discussions.

\begin{figure}[t]
    \centering
    \includegraphics[width=0.45\textwidth]{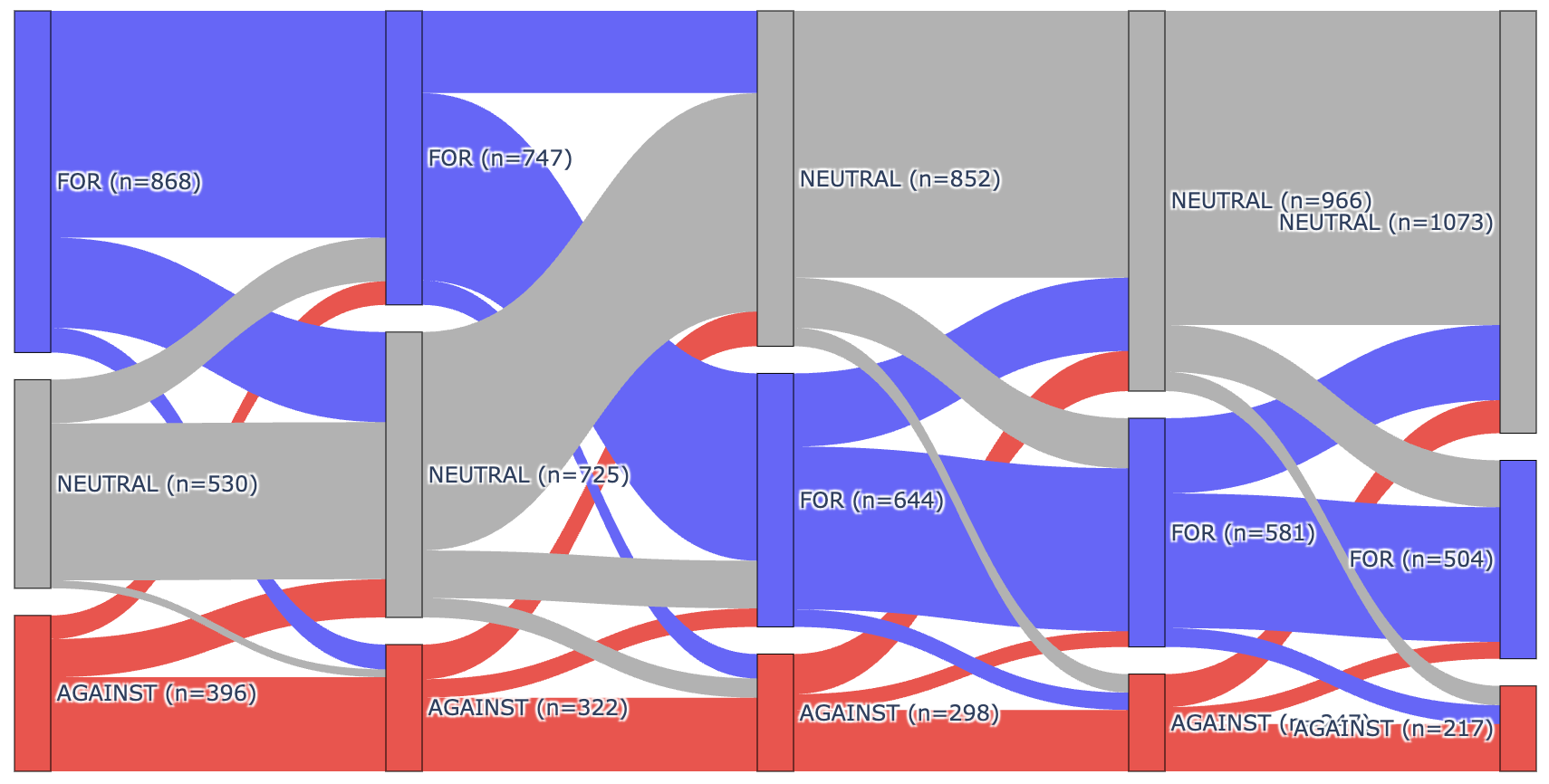}
    \caption{Shifts in user stance over 5 turns in a conversation.}
    \label{fig:sankey}
\end{figure}

\section{Value of the Data and Possible Applications}
To facilitate reuse and reproducibility, we release the dataset as three separate CSV files: \texttt{new\_truths-all.csv}, \texttt{new\_authors-all.csv}, and \texttt{ground\_truth.csv}. The TruthStance dataset is publicly available via Zenodo \cite{Ameen2026TruthStance} under a Creative Commons Attribution--NonCommercial 4.0 International (CC BY 4.0) license, which allows anyone to redistribute its contents. To ensure provenance and consistent comparison across studies, the dataset described in this paper is released as a static versioned snapshot.

TruthStance is designed to support responsible research on conversational dynamics and argumentation on Truth Social. In addition, we commit to maintaining public documentation and issue tracking for the dataset (e.g., schema clarifications, bug fixes in preprocessing scripts, and best practices for use) via our GitHub repository. TruthStance follows the FAIR principles: it is \textit{Findable} through Zenodo with a persistent identifier, \textit{Accessible} via direct download through the repository interface, \textit{Interoperable} due to its standardized CSV format, and \textit{Reusable} through clear licensing and comprehensive documentation.

Potential applications of TruthStance include: (i) benchmarking argument mining and stance detection models in platform-specific conversational settings, (ii) studying disagreement trajectories and stance propagation in comment threads, (iii) analyzing how affective signals (e.g., sentiment and toxicity) correlate with argumentative language, (iv) modeling engagement patterns and structural properties of conversation threads, and (v) supporting research on conversation moderation, polarization, and online deliberation (when used responsibly and with appropriate safeguards).

\section{Limitations}
TruthStance has several limitations that should be considered when interpreting results or reusing the dataset. First, the dataset is collected from a single platform and therefore reflects platform-specific norms, affordances, and user populations. As a result, findings derived from TruthStance should not be interpreted as representative of broader public opinion or general online discourse.

Second, the dataset is subject to selection effects introduced by data collection and filtering. We filtered out posts with fewer than 3 comments, which introduces selection bias toward high-engagement posts; stance dynamics in low-comment threads may differ. We were also only able to scrape comments from the posts included in an existing dataset, so any selection bias introduced there would propagate here as well. In addition, the dataset captures only publicly available content at the time of collection; deleted posts, removed accounts, and private interactions are not observed.

Third, while we provide high-quality human annotations for argument mining and stance detection, it is not feasible to label the full comment corpus due to time and cost constraints. Consequently, model training and evaluation rely on a representative annotated subset, and performance estimates may vary across topics, user communities, and conversation threads not fully covered by the labeled data.

Finally, stance is inherently contextual and can be ambiguous, especially in short replies, sarcasm, humor, or cases involving multiple targets. Although our stance labels are defined relative to the immediate parent comment, inferring stance relative to the original post requires a composition assumption along comment paths. In particular, when a neutral intermediate comment occurs, stance propagation becomes undefined; therefore, OP-level stance inference is conservatively truncated for descendant branches originating from neutral comments.

\section{Discussion of Potential Misuse}
Because TruthStance is derived from social media content, we consider privacy and downstream harms as central ethical concerns. Our collection procedure uses only publicly accessible posts and comments, and we follow a data-minimization approach by retaining only the information necessary for analysis. We avoid reporting specific usernames in the paper and focus on aggregate statistics and trends. During development, the dataset was stored and processed securely, with access restricted to authorized researchers to reduce the risk of unauthorized disclosure.

Despite these safeguards, there remains a risk that the dataset or derived models could be misused. For example, stance and toxicity signals could be applied to enable political profiling, targeted harassment, or automated labeling of individuals or communities based on their online expression. Such uses may reinforce harmful stereotypes or be deployed for discriminatory or punitive purposes. We therefore emphasize that TruthStance is intended for non-commercial research and should be used responsibly, with careful attention to context, uncertainty in model predictions, and the potential for bias. We encourage researchers to avoid individual-level interpretation, to report findings at the group or population level, and to apply appropriate ethical review procedures when extending this work.

\section{Conclusion}
We introduce TruthStance, a publicly available dataset of conversational threads from Truth Social, partially annotated for argument presence and stance in conversation threads. In addition to releasing the dataset and documentation, we provide descriptive analyses of engagement, conversational structure, and stance distributions across topics, time, and users. Our results suggest that content-based signals, particularly toxicity and sentiment, are strongly associated with argument labeling, and that stance dynamics in conversation threads exhibit measurable patterns of agreement and disagreement. We hope TruthStance supports future work on argument mining, stance detection, and the study of polarization and conversational dynamics in online communities, while encouraging responsible and privacy-aware use of these data.


\bibliography{aaai25}

\appendix
\section{A: Annotation Guidelines}
\label{app:annotations}
Here we provide the full annotation instructions used for both tasks.

\begin{enumerate}
    \item General Instructions: Annotations should be based solely on the content of the post and comment under consideration. When determining stance, focus on the substantive position expressed rather than tone, sarcasm, or politeness markers. In cases of ambiguity, annotators should select the label that best reflects the overall argumentative intent of the comment.

    \item Argument Mining: Annotate a post as \textbf{Argumentative} if it contains both a claim and at least one supporting premise. Posts lacking either component should be labeled \textbf{Non-Argumentative}.
    \begin{itemize}
      \setlength{\itemsep}{0.2em}
      \item \textbf{Claim}: The main point or position the author wants readers to accept.
      \item \textbf{Premise}: A statement offered as support or justification for the claim. Implicit premises count if they clearly support the claim.
    \end{itemize}
    
    \item Stance Detection: For each comment in response to its parent comment (or post), assign one of the following stance labels relative to the claim advanced in the parent:
    \begin{itemize}
      \setlength{\itemsep}{0.2em}
      \item \textbf{FOR}: The comment clearly supports or agrees with the tweet’s claim or premises.
      \item \textbf{AGAINST}: The comment clearly opposes, challenges, or rejects the tweet’s claim or premises.
      \item \textbf{NEUTRAL}: The comment does not clearly support or oppose the tweet’s claim or premises, or the tweet expresses no clear claim. This includes replies that are irrelevant, vague, purely expressive, off-topic, or promotional.
    \end{itemize}
\end{enumerate}

\section{B: Prompt Templates}
\label{app:prompts}
\paragraph{Few-Shot (FS).}
\noindent\emph{Instruction:} Ignore advertisements, tone, language quality, and factual accuracy. For each tweet, return your annotation in exactly the following format:
\begin{quote}\ttfamily\small
[id: <id>, annotation: (argumentative/not argumentative)]\\
tweets: [..]\\
output: [...]
\end{quote}

\paragraph{Chain-of-Thought (CoT).}
\noindent\emph{Instruction:} Ignore advertisements, tone, language quality, and factual accuracy. For each tweet, reason step-by-step before returning your annotation in exactly the following format:
\begin{quote}\ttfamily\small
[id: <id>, annotation: The response is (argumentative/not argumentative) because: (brief justification)]
\end{quote}

\paragraph{Few-Shot + Chain-of-Thought (FSCoT).}
\noindent\emph{Instruction:} Ignore advertisements, tone, language quality, and factual accuracy. For each tweet, reason step-by-step before returning your annotation in exactly the following format:
\begin{quote}\ttfamily\small
[id: <id>, annotation: The response is (argumentative/not argumentative) because: (brief justification)]\\
tweets: [..]\\
output: [...]
\end{quote}

\end{document}